\documentclass{article}

\usepackage{arxiv}
\usepackage{cite}
\usepackage{amsmath,amssymb,amsfonts}
\usepackage{algorithmic}
\usepackage{graphicx}
\usepackage{subfigure}
\usepackage{textcomp}
\usepackage{xcolor}
\usepackage{booktabs}
\def\BibTeX{{\rm B\kern-.05em{\sc i\kern-.025em b}\kern-.08em
    T\kern-.1667em\lower.7ex\hbox{E}\kern-.125emX}}
\usepackage[draft]{hyperref}
\definecolor{darkred}{rgb}{0.85,0,0}
\definecolor{darkgreen}{rgb}{0,0.6,0}
\definecolor{darkblue}{rgb}{0,0,0.5}
\hypersetup{hidelinks, breaklinks, colorlinks,bookmarks=false,
  linkcolor={darkblue}, citecolor={darkblue}, urlcolor={darkblue}}

\usepackage[pscoord]{eso-pic}

\newcommand{\OO}{\mathcal{O}}
\begin{document}

\title{HeunNet: Extending ResNet using Heun's Methods}

\author{ \href{https://orcid.org/0000-0003-1020-3322}{\includegraphics[scale=0.06]{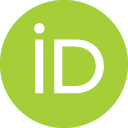}\hspace{1mm}Mehrdad Maleki} \\
	Dept of Computer Science\\
		Maynooth University\\
	Maynooth, Ireland \\
	\thanks{Mehrdad Maleki funded by Irish Research Council (IRC) Fellowship GOIPD/2019/803.}
	\And
	\href{https://orcid.org/0000-0001-9051-1370}{\includegraphics[scale=0.06]{orcid.png}\hspace{1mm}Mansura Habiba} \\
	Dept of Computer Science\\
	Maynooth University\\
	Maynooth, Ireland \\
	\And
	\href{https://orcid.org/0000-0003-0521-4553}{\includegraphics[scale=0.06]{orcid.png}\hspace{1mm}Barak A. Pearlmutter} \\
	Department of Computer Science \& Hamilton Institute\\
	Maynooth University\\
	Maynooth, Ireland \\
}

\maketitle

\begin{abstract}
   There is an analogy between the ResNet (Residual Network) architecture for deep neural networks and an Euler solver for an ODE. The transformation performed by each layer resembles an Euler step in solving an ODE. We consider the Heun Method, which involves a single predictor-corrector cycle, and complete the analogy, building a predictor-corrector variant of ResNet, which we call a HeunNet. Just as Heun's method is more accurate than Euler's, experiments show that HeunNet achieves high accuracy with low computational (both training and test) time compared to both vanilla recurrent neural networks and other ResNet variants.
\end{abstract}

\section{Introduction}


A ResNet or Residual Network \cite{resnet} can be viewed as using Euler's Method to integrate a time-dependent ordinary differential equation. Since Euler's Method is very inaccurate, with $\text{error} = \OO(\Delta t)$, one might conjecture that some of the power of a ResNet with many layers is being devoted to compensating for integration error introduced by this crude approximation. This suggests the use of a more accurate integration method, which might result in less error being introduced, and therefore derive greater benefit from the same number of layers.

Here we use the Heun Method \cite{heun}, which is much more accurate: $\text{error} = \OO(\Delta t^2)$. In our HeunNet model, each block computes $x_t$, the cell state at time~$t$, along with an initial estimate of the next cell state,~$x_{t+1}$.  The Heun Method finds $x_t$ using both $x_{t+1}$ and an initial estimate of~$x_t$. This approach helps proposed HeunNet model to provide better result with low memory footprint. 




This Heun Method approach can be adapted for a wide variety of layered neural network architectures. HeunNet model only takes input for each state. It then computes both the current and the following cell states using the Heun Method, as shown in Eq.~\eqref{eq:heun}.

Each block of the HeunNet neural network uses the same weight and computation method. Therefore, it does not cause additional memory or computation energy. Although each iteration can take longer than simple neural networks, the result shows that the proposed neural network achieves higher accuracy than other neural networks at an early stage of training.

The main objectives of this work are:
\begin{itemize}
	\item Design a new neural network architecture based on the Heun Method (HeunNet).
	\item Investigate different  evaluation metrics to assess the performance of the proposed model.
\end{itemize}

\section{Heun's Method for solving ODEs}
\begin{figure*}[t]
  \includegraphics[trim={0 0 10 10},clip,width=0.45\columnwidth]{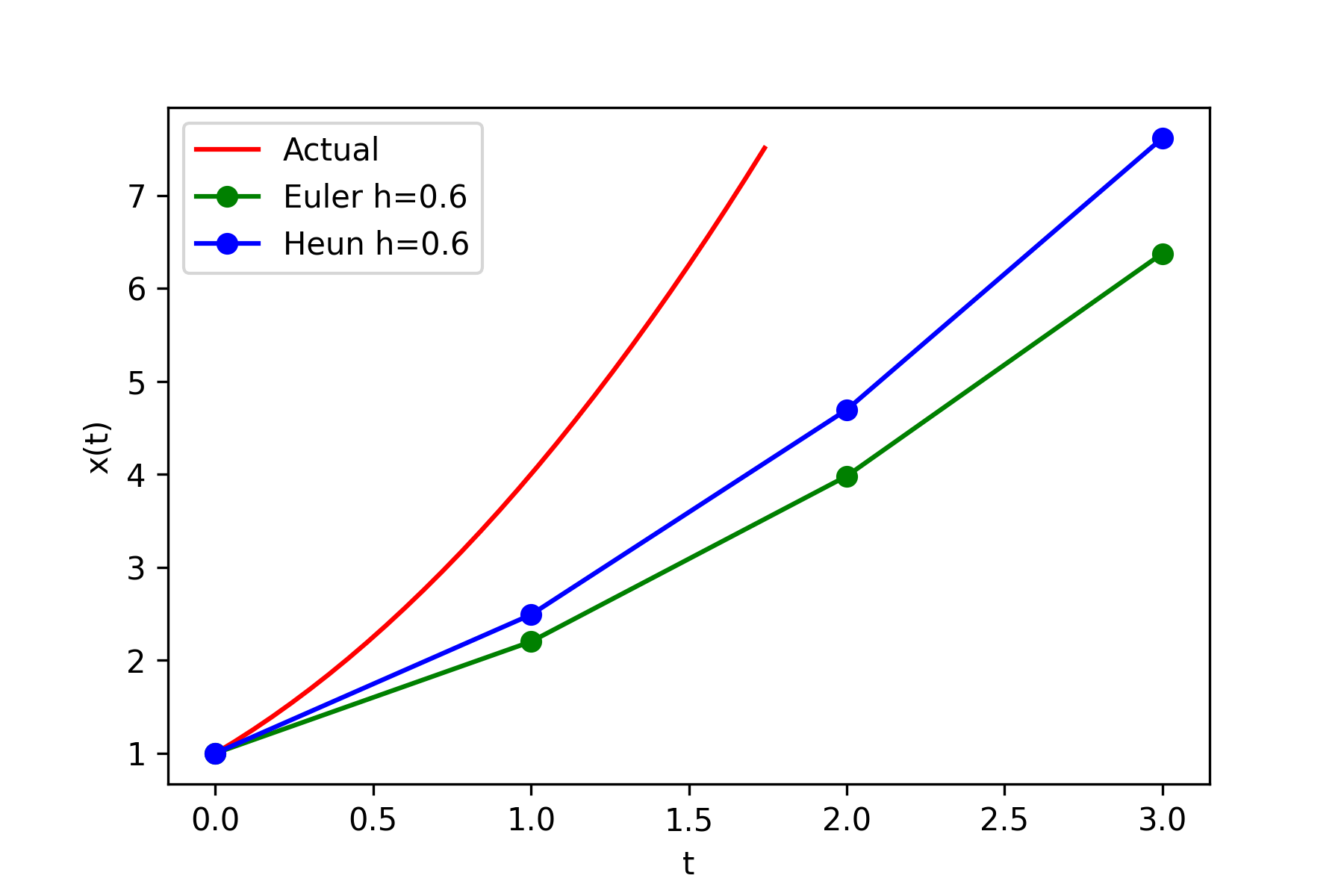}
  \hfill
  \includegraphics[trim={0 0 10 10},clip,width=0.45\columnwidth]{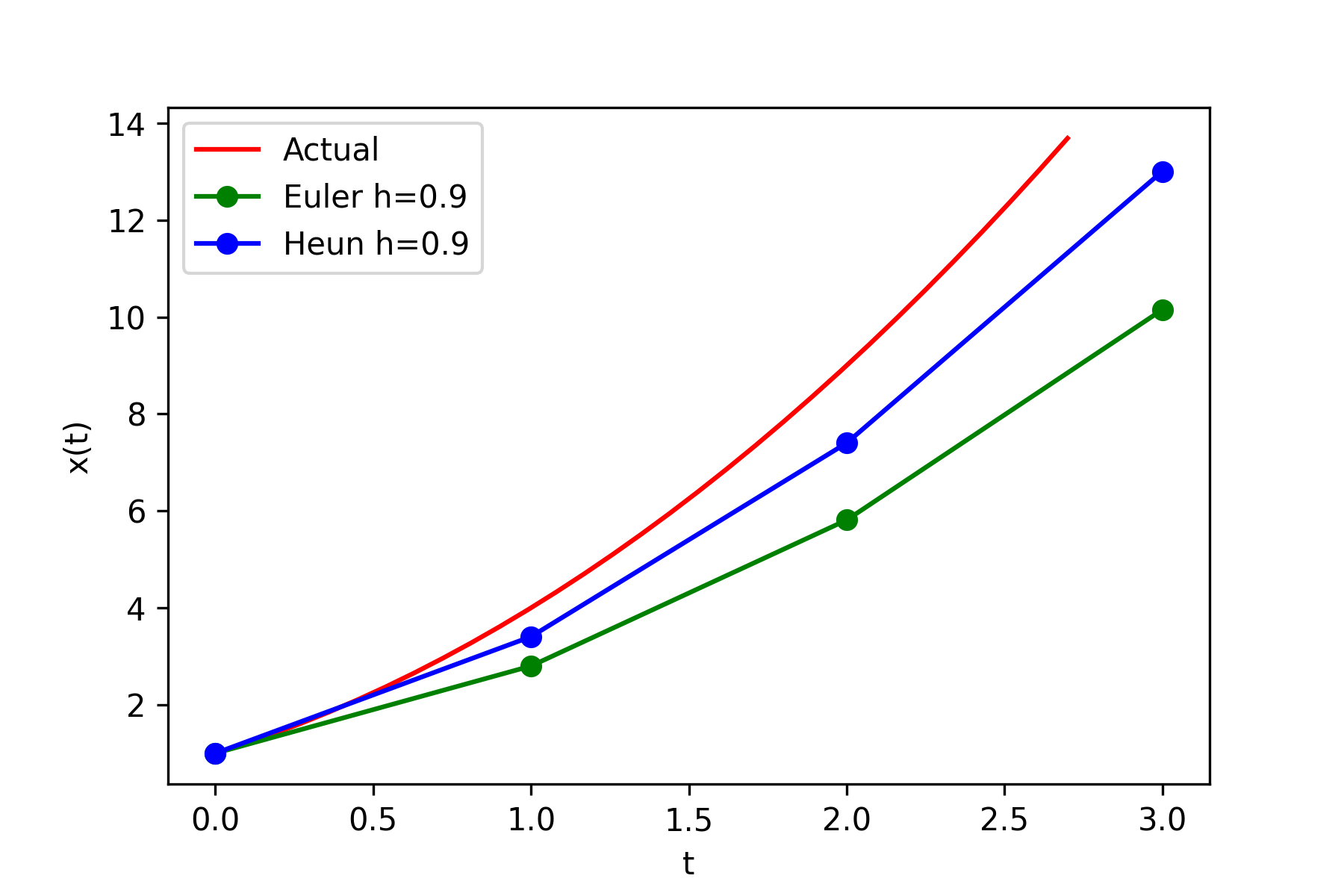}
  \caption{Euler and Heun Method for $\dot{x}=2\sqrt{x}$ with $x(0)=1$ and $h=0.6$ (left) and $h=0.9$ (right). The analytic solution of this ODE is $x(t)=(t+1)^2$ (red curve).}
  \label{fig:euler_heun}
\end{figure*}
Heun's Method is a numerical method for solving ordinary differential equations (ODEs). Heun's Method is sometimes referred to as an ``improved Euler'' method, but it is better viewed as similar to a two-stage Runge-Kutta method \cite{heun}. For solving an ODE $\dot{x}=f(t,x)$ with initial condition $x(0)=x_0$ using Euler's method, we use the slope $f$ at $x_0$ and move along the tangent line to $x$ at $x_0$ to reach to estimate $x_{t+1}$,
\begin{equation}
x_{t+1}=x_t+hf(t,x_t)
\label{eq:euler}
\end{equation}
where $h$ is the step size. If the actual solution is convex (or concave), the Euler method will underestimate (or overestimate) the next state of the system. So in the long term, the numerical solution diverges from the correct solution. A better estimate can be obtained using an estimate of the slope at the next step and using the average of this slope with the current step's slope. This is the essence of Heun's Method,
\begin{subequations}
\begin{align}
\tilde{x}_{t+1}&=x_t+hf(t,x_t) \label{eq:h_predict} \\
x_{t+1}&=x_t+h\frac{f(t,x_t)+f(t+1,\tilde{x}_{t+1})}{2}
 \label{eq:h_correct}
\end{align}
\end{subequations}
In predictor-corrector methods like this there are two steps, ``prediction'' and ``correction.'' A ``predictor'' \eqref{eq:h_predict} estimates the value at the next step, and this estimate is improved by a ``corrector'' \eqref{eq:h_correct}.

The difference between Euler and Heun Method for the ODE is $\dot{x}=2\sqrt{x}$ and initial condition $x(0)=1$ for $h=0.6$ and $h=0.9$ plotted in Fig.~\ref{fig:euler_heun}.

\section{Motivation}

Consider a neural network with $L$ layers such that the number of neurons in all the layers are equal.  Let $X_0$ be the input of the network and let $W_0$ be the square matrix (since all layers have the same number of neurons) of weight between the input and the first layers. Let $Z_{1}=W_0 X_0$ and $X_{1}=\sigma(Z_{1})$. So we define $\mathcal{F}(X_k)=\sigma(W_kX_k)=X_{k+1}$. The regular architecture of a neural network is straightforward as shown in Eq.~\eqref{eq:regular_nn}:

\begin{equation}
X_{k+1}=\mathcal{F}(X_k)
\label{eq:regular_nn}
\end{equation}
A residual neural network (ResNet) uses skip connections to deal with the problem of vanishing gradients during the training.
\begin{equation}
	X_{k+1}=X_k+\mathcal{F}(X_k)
	\label{eq:resnet}
\end{equation}
This architecture is quite efficient when dealing with an intense neural network, even in combination with convolutional layers.
However, if we look closely, we realize that this is similar to Euler's method for solving an ODE with step size $h=1$ \cite{neuralode}. The motivation is that ResNet is a modification of the Euler method, so HeunNet is the modification of Heun's Method.

\section{Proposed Model: HeunNet Model}
Our motivation inspired by the Heun Method \cite{heun}. Since ResNet \cite{resnet} is a discretization of the Euler method, we could see the Heun Method as the extension of ResNet.
In a nutshell,\\[0.5ex]
\centerline{\fbox{ResNet : Euler's Method :: HeunNet : Heun's Method}}

\begin{figure}[t]
  \centerline{\includegraphics[trim=20 150 15 5,clip,width=\columnwidth]{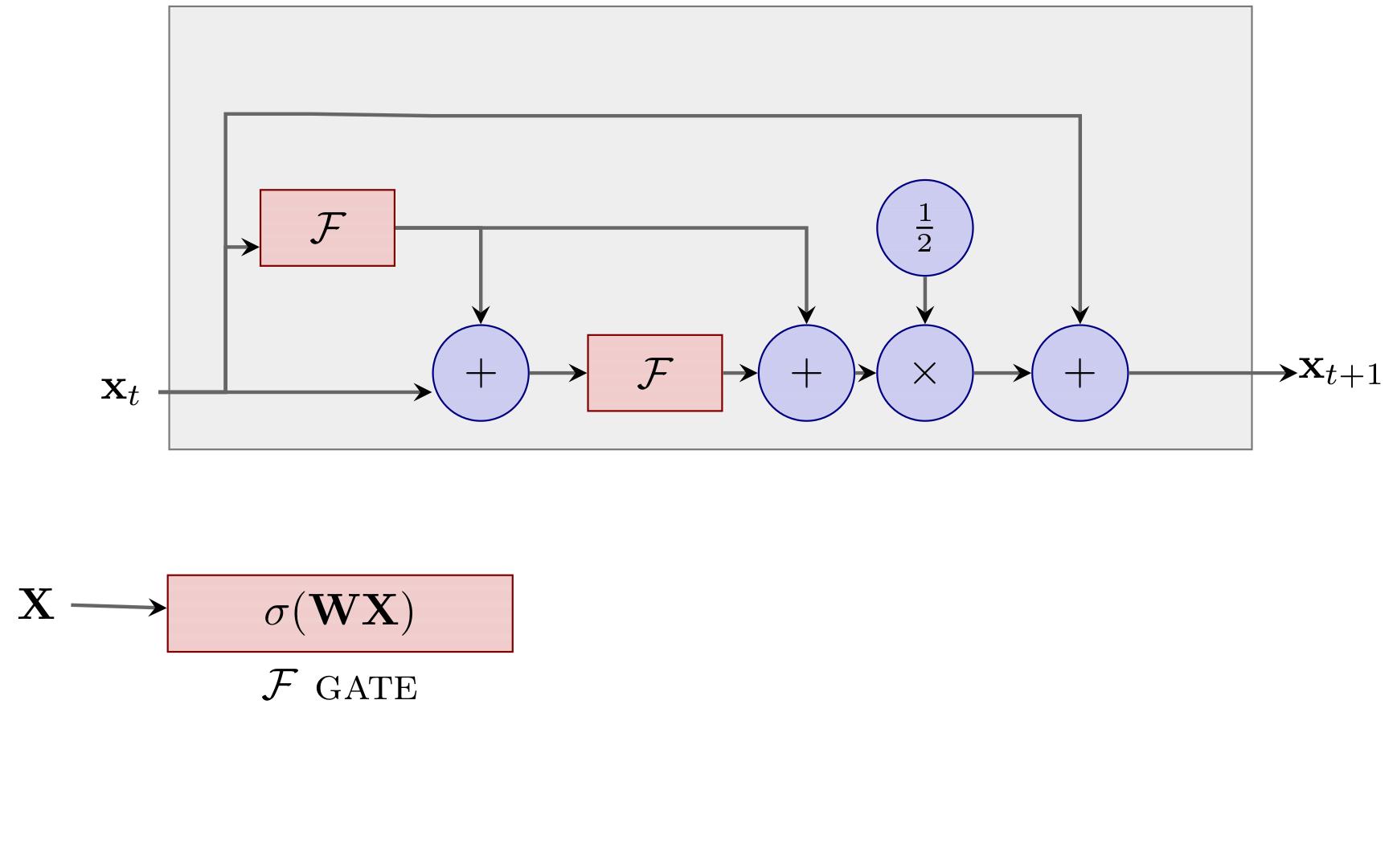}}
  \caption{The architecture of proposed Heun based Recurrent Neural Network (HeunNet)}
  \label{fig:architecture}
\end{figure}

We propose another architecture for neural networks motivated by the Heun Method for solving ODE. In our model, the output of the next layer is a combination of the output from ResNet and the output of the previous layer as shown in Eq.~\eqref{eq:predictor}.
\begin{subequations}\label{eq:heun}
  \begin{align}
\label{eq:predictor}
\tilde{X}_{k+1}&=X_k+\mathcal{F}(X_k)
\\
\label{eq:corrector}
X_{k+1}&=X_k+\tfrac{1}{2}(\mathcal{F}(X_k)+\mathcal{F}(\tilde{X}_{k+1}))
  \end{align}
\end{subequations}
Fig.~\ref{fig:architecture} shows the essence of this architecture.

Residual Neural Network is a discretization of the Euler method. In the residual network, the state of the layer $k+1$ is related to the state of the layer $k$ as shown in \eqref{eq:resnet}, where $\mathcal{F}$ can be an arbitrary layer transition function.
But if we use Heun's Method for solving $\dot{x}(t)=\mathcal{F}(x(t))$ with initial condition $x(0)=x_0$ and let step size be $h=1$, we obtain the architecture of the Heun Neural Network (HeunNet), i.e., Eq.~\eqref{eq:heun}.

The Heun Method has quadratic accuracy, while the accuracy of Euler's method is linear. For this reason, the Heun Method is generally preferred to Euler's method when numerically solving ODEs. A similar situation is occurring for the Heun Neural Network (HeunNet), which appears to converge to a more accurate result in fewer layers.

\subsection{Gradient Propagation in a HeunNet}

The original motivation for ResNet was to avoid the vanishing gradient problem by causing the Jacobian of the transition between layer $k$, and layer $k+1$ to be a near-identity matrix.
Both the ResNet equation \eqref{eq:resnet} and the analogous HeunNet equation \eqref{eq:heun} are of the form $X_{k+1} = X_k + \textit{something}$, which accomplishes this.
Like ResNet, HeunNet is an entirely feedforward computational process that can be expressed using standard numeric linear algebra routines. As such, it is straightforward to implement in any deep learning framework, such as PyTorch. The gradient can be automatically calculated via backpropagation, i.e., reverse mode Automatic Differentiation \cite{baydin2018automatic}, and that process is just as automatic as for ResNet.

Despite its conceptual similarity, and as we see below, numerical simulations show that HeunNet obtains better performance than ResNet across various tasks.

\section{Extension of HeunNet Model}

In the Heun's Method for solving ODEs, if we put more weights on the corrector's second term we will have a better approximation of the actual function. In the Heun's Method, we first compute the $f(x_k)$ as the slope of the current state and $f(\tilde{x}_{k+1})$ as the approximation coming from Euler method and get the average of these two as the corrector. But if we put more weights on the $f(\tilde{x}_{k+1})$ and use convex hull of $f(x_k)$ and $f(\tilde{x}_{k+1})$, i.e., $(1-\alpha)f(x_k)+\alpha f(\tilde{x}_{k+1})$ with $\alpha$ near to~1, then we have better approximation than Heun's Method:
\begin{align*}
\tilde{x}_{k+1}&=x_k+h\,f(x_k)\\
x_{k+1}&=x_k+h\,((1-\alpha)f(x_k)+\alpha f(\tilde{x}_{k+1}))
\end{align*}
where $0\leq \alpha \leq 1$ and $h$ is the step size. Fig.~\ref{fig:euler_heun} shows the difference between Heun's Method and extended Heun's Method for $\dot{x}=2\sqrt{x}$ with $x(0)=1$.

For $h=1$ we obtain the architecture of a Neural Network which is an extension of HeunNet (ExtendedHeunNet), i.e.,
\begin{align}
\tilde{x}_{k+1}&=x_k+f(x_k)\\
x_{k+1}&=x_k+(1-\alpha)f(x_k)+\alpha f(\tilde{x}_{k+1})
\end{align}

\begin{figure}[t!]
  \centerline{\includegraphics[trim=20 150 15 5,clip,width=0.9\columnwidth]{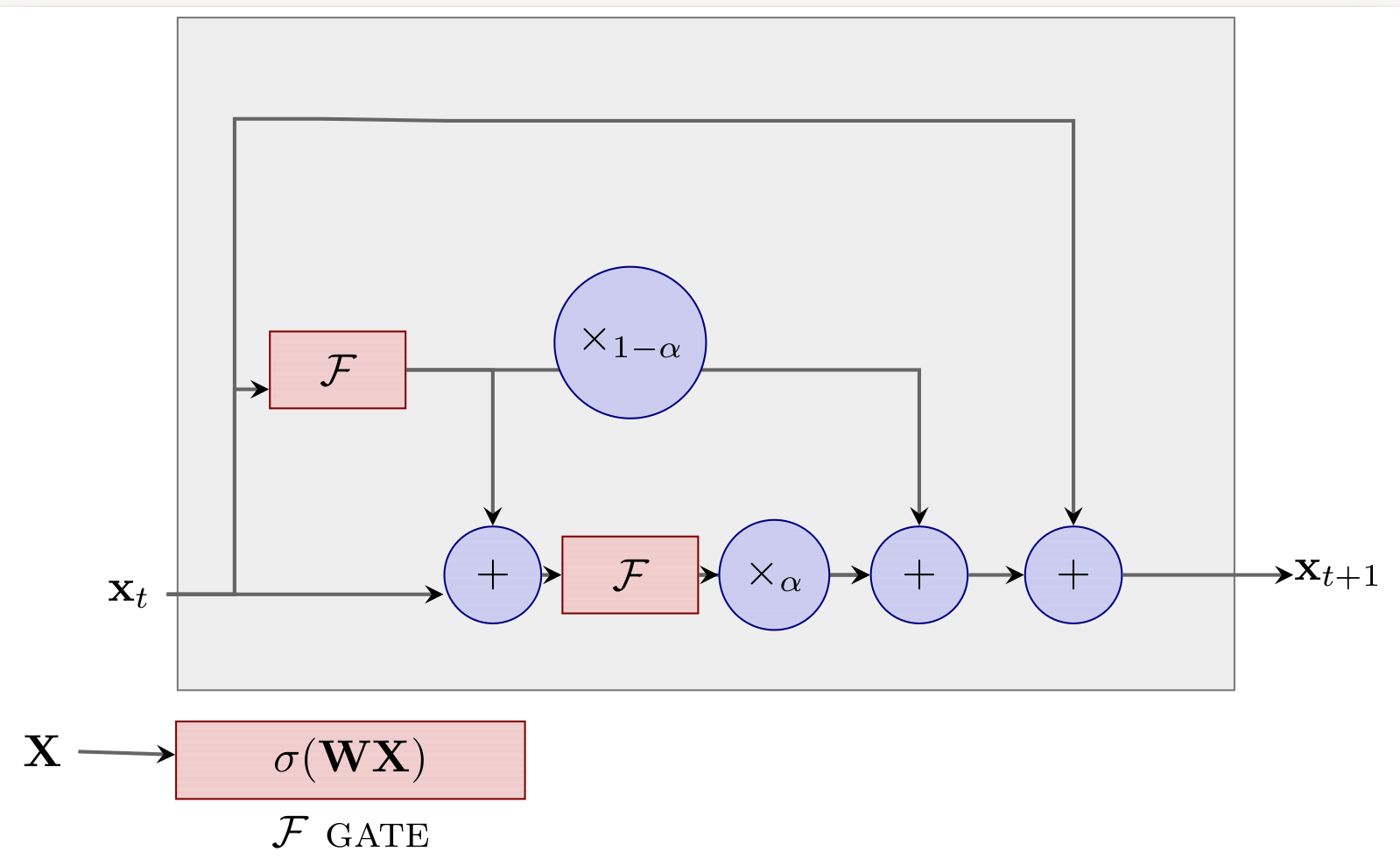}}
  \caption{The architecture of proposed Extended Heun based Recurrent Neural Network (ExtendedHeunNet)}
  \label{fig:exhnn}
\end{figure}

\begin{itemize}
\item If $\alpha=0$ then $x_{k+1}=\tilde{x}_{k+1}$ and we have \textbf{ResNet}.
\smallskip
\item If $\alpha=\frac{1}{2}$ then we have \textbf{HeunNet}.
\smallskip
\item If $\alpha\neq \frac{1}{2}$ then we have \textbf{ExtendedHeunNet}.
\end{itemize}
The question is, which values of $\alpha$ gives us better approximation? Does this modification improves the performance of HeunNet? Since we need more weights on the next state to have a better approximation of the actual function, we choose $\alpha$ to be close to 1. The architecture of this extended Neural Network is shown in Fig.~\ref{fig:exhnn}.

\section{Results}
For performance evaluation of proposed new model, we have chosen two tasks, (i) MNIST dataset classification \cite{mnist} and (ii) ECG classification \cite{ecg}. We compared the HeunNet model with other Recurrent Neural Network, e.g., LSTM \cite{lstm} and GRU \cite{gru}.

\begin{figure*}[t!]
  \includegraphics[trim=5 5 5 5,clip,width=0.48\textwidth]{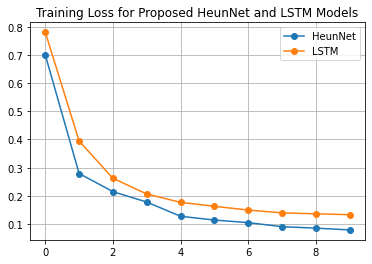}
  \hfill
  \includegraphics[trim=5 5 5 5,clip,width=0.48\textwidth]{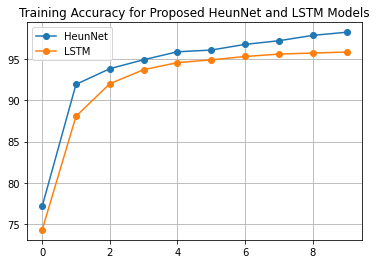}
  \caption{HeunNet and LSTM model loss (left) and accuracy (right) for 10 iterations on MNIST classification task}
  \label{fig:mnist}
  \vspace{4ex}
  \includegraphics[trim=5 5 5 5,clip,width=0.49\textwidth]{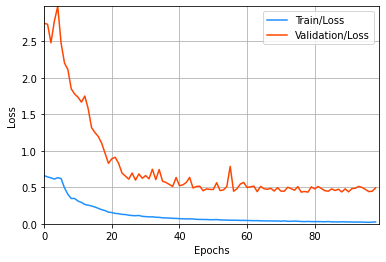}
  \hfill
  \includegraphics[trim=5 5 5 5,clip,width=0.49\textwidth]{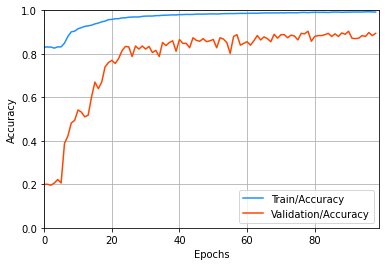}
  \caption{LSTM model Loss (left) and accuracy (right) for 100 iterations on ECG heartbeat classification task}
  \label{fig:ecg_task}
  \vspace{4ex}
  \includegraphics[trim=5 5 5 5,clip,width=0.49\textwidth]{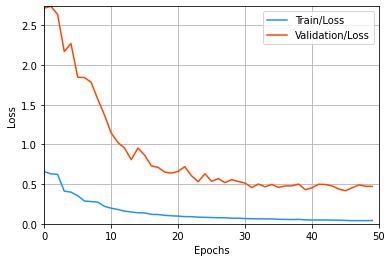}
  \hfill
  \includegraphics[trim=5 5 5 5,clip,width=0.49\textwidth]{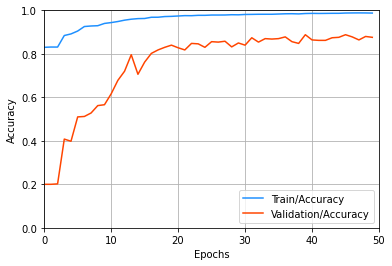}
  \caption{HeunNet model loss (left) and accuracy (right) for 50 iterations on ECG heartbeat classification task}
  \label{fig:ecg_task_heun}
\end{figure*}

\subsection{MNIST classification Task}

Fig.~\ref{fig:mnist} shows the comparative analysis of loss and accuracy for LSTM and HeunNet models over ten iterations. LSTM achieves an accuracy of 95.44\% after the 10th iteration. On the other hand,  the HeunNet model achieves an accuracy of 98.26\% at the 10th iteration.

\subsection{ECG heartbeat classification Task}
For this task, we used the dataset for MIT-BIH Arrhythmia Database v1.0.0  PhysioNet \cite{mitbih}. Table~\ref{tab:ecg_datset} shows some characteristics of the used datasets. This dataset consists of signals that correspond to electrocardiogram (ECG) shapes of heartbeats for the normal case and shapes caused by arrhythmias and other myocardial infarctions.  There are five categories of beats present in the dataset as Nonectopic beat (N), Supraventricular ectopic beat (S), Ventricular ectopic beat (V),  Fusion beat (F) and Unknown beat (Q).

\begin{table}[bt]
  \centering
  \caption{Parameters of Dataset Used}
  \label{tab:ecg_datset}
  \begin{tabular}{lll}
    \toprule
    \textbf{\#} & \textbf{Parameter}  & \textbf{Value} \\\midrule
    1 & Number of Samples    & 109446     \\
    2 & Number of Categories & 5          \\
    3 & Sampling Frequency   & 125 Hz     \\
    4 & Classes              & [N: 0, S: 1, V: 2, F: 3, Q: 4] \\\bottomrule
  \end{tabular}
\end{table}

Fig.~\ref{fig:ecg_task} shows the loss and accuracy for LSTM model over 100 iterations.  The best iteration for the LSTM model in the ECG heartbeat classification task is 79 with an accuracy  90.40\%.

With the HeunNet model, we get the best accuracy of 98.80\%. Fig.~\ref{fig:ecg_task} shows the loss and accuracy for LSTM model for only 50 iterations. 

The proposed model can achieve higher accuracy within only half of the number of iterations required for other recurrent neural networks.

\subsection{Time Series prediction}

\begin{figure}[t!]
\centering
  \includegraphics[trim=5 5 5 5,clip,width=0.65\columnwidth]{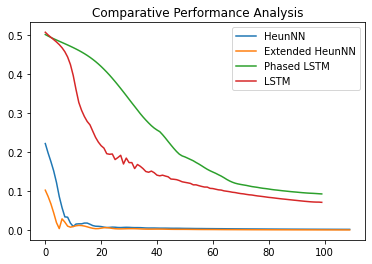}
  \caption{Loss for LSTM, Phased-LSTM, and HeunNet models over 100 iterations for Sine Wave generation}
  \label{fig:taskc}
\end{figure}

As shown in \eqref{eq:heun}, the neural network $\mathcal{F}$ can be any neural network. For time series generation task, we used a simple LSTM neural network for $\mathcal{F}$ in  \eqref{eq:heun}. This Heun neural network-based LSTM model is compared against Phased LSTM \cite{phasedlstm} and vanilla LSTM model. A sine wave of length $16\pi$ is used for the dataset for this task. Each model is trained for 100 iterations. Fig.~\ref{fig:taskc} shows the comparative performance analysis for three models. As shown in Fig.~\ref{fig:taskc}, the HeunNet model outperformed the other two model.

Proposed ExtendedHeunNet model's prediction is more accurate than other HeunNet and also other two neural network models. ExtendedHeunNet provides an accuracy of 97.30\%  with $\alpha =0.8$ in just 2 iterations for Task C. For ExtendedHeunNet,  $0.75 < \alpha \le 0.8$ provides best accuracy. If $\alpha$ move towards 1, the accuracy starts to drop. 

\begin{figure*} [http]
	\subfigure[LSTM model trained for 100 iterations]{\includegraphics[width=0.5\textwidth]{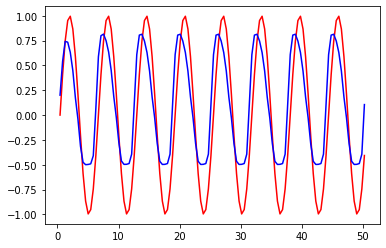}}
	\hfill
	\subfigure[Phased-LSTM model trained for 100 iterations]{\includegraphics[width=0.5\textwidth]{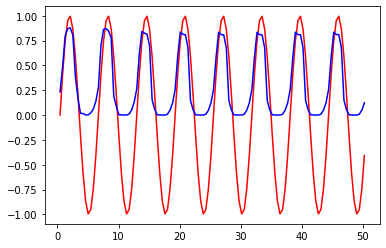}}
	\vfill
	\subfigure[HeunNet model trained for 25 iterations]{\includegraphics[width=0.5\textwidth]{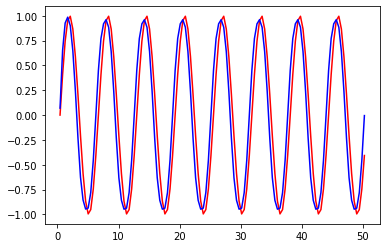}}
	\hfill
	\subfigure[ExtendedHeunNet model trained for 2 iterations with $\alpha=0.8$]{\includegraphics[width=0.5\textwidth]{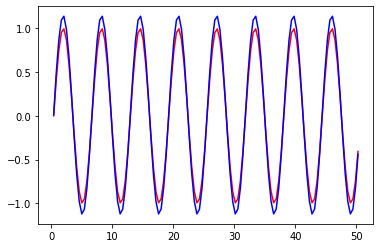}}
	\vfill
		\subfigure[ExtendedHeunNet model trained for 10 iterations $\alpha=0.8$]{\includegraphics[width=0.5\textwidth]{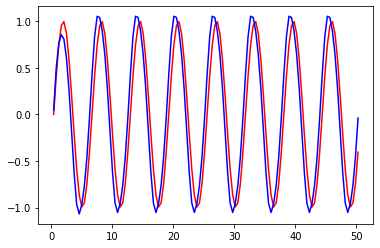}}
\caption{Training different models to generate a sine wave.}
\label{fig:taskc2}
\end{figure*}

Fig.~\ref{fig:taskc2} shows generated sine wave for LSTM, Phased-LSTM and HeunNet  model. The HeunNet and ExtendedHeunNet model's prediction are more accurate than the predictions of the other two models.

\subsection{Performance Evaluation}
Typically, the accuracy of RNN models (LSTM or GRU) increase with the increase in the number of iteration. Therefore, the training time is comparatively long. We experimented with different ranges of iteration for different problems. For task A, within ten iterations, LSTM achieve 95.44\% accuracy, and GRU achieves 92.01\% accuracy. However, for the same number of iteration, the  HeunNet model achieves 97.98\% accuracy. The result shows that the proposed model needs less training time to achieve high accuracy.

In a traditional bi-directional vanilla recurrent neural network, corresponding parameters are combined into a single LSTM. On the other hand, in this proposed model, the results of two individual neural networks are combined using the Heun Method \eqref{eq:heun}.

In this section, both classification and time series prediction tasks are evaluated for different recurrent models against the HeunNet model. We can conclude that the proposed model can achieve higher accuracy with almost half the number of iterations than other models. The HeunNet model does not need to train for a long time to get the optimized result.

\section{Discussion}
ResNet has the form of an Euler method for discretizing and solving ODEs, and HeunNet and its extensions can be similarly viewed as having the form of Heun's Method. Some other groups have independently explored similar ideas for extending ResNet based on other numeric ODE solvers. In particular, MomentumNets \cite{sander2021momentum} are constructed using an analogy with a numeric solver for second-order ODEs, while our HeunNet and its extension are derived from solution methods for first-order ODEs. Since second-order ODEs can be reduced to first-order ODEs using standard techniques, it should be possible to put these all into a common framework. MomentumNets have ResNet as a special case, but our extended HeunNet has both ResNet and HeunNet as special cases. And finally, in MomentumNets the velocity term is defined as a function of the current state in the first stage, and in the second stage this term is combined with the input of the previous layer, while our HeunNet or its extension use ResNet as the approximation for the next state and the second term is a correction for the error introduced in the first step. So the HeunNet first makes an approximation for the next state, and a weighted average of this approximation and its correction constitutes the space of candidates for the next state of the system.

Heun's Method falls in the general class of predictor-corrector methods: it makes an initial guess as to the correct state of the next time step, then corrects that estimate using more accurate information based on this first estimate.
This can be viewed as a local causality violation, as it makes the update between layers $t$ and $t+1$ depend not just on the state of layer $t$, but also on that of layer $t+1$. However the method remains causal, unlike time-series processing techniques like a Bi-directional RNN, which consists of a forward-in-time pass along with a corresponding backward-in-time pass.
The technique proposed here might be naturally extended to that setting by making each of these passes update using a Heun's Method step rather than an Euler step. So we should not view these as in competition, but rather as potentially synergistic.

\section{Conclusion}
We proposed a new neural network model that leverages Heun's Method for hidden state optimization in this work. We also developed this architecture that includes HeunNet as a special case. Performance evaluation demonstrates that both proposed models outperform existing recurrent neural networks in case of solving different problems. The proposed models also takes less time for training and can achieve higher accuracy. This architecture leverage the strength of Heun's Method that ensures higher accuracy. The error generated by the first step of the Heun Method is compensated by the corrector, i.e., the second term, and reduce the errors \cite{heun}.

\bibliographystyle{plain}
\bibliography{heunnet_Arxiv} 

\begin{thebibliography}{10}

\bibitem{baydin2018automatic}
Atılım~Güneş Baydin, Barak~A. Pearlmutter, Alexey~Andreyevich Radul, and
  Jeffrey~Mark Siskind.
\newblock Automatic differentiation in machine learning: a survey.
\newblock {\em Journal of machine learning research}, 18, 2018.

\bibitem{neuralode}
Ricky T.~Q. Chen, Yulia Rubanova, Jesse Bettencourt, and David Duvenaud.
\newblock Neural ordinary differential equations.
\newblock {\em arXiv preprint arXiv:1806.07366}, 2018.

\bibitem{gru}
Kyunghyun Cho, Bart Van~Merri{\"e}nboer, Caglar Gulcehre, Dzmitry Bahdanau,
  Fethi Bougares, Holger Schwenk, and Yoshua Bengio.
\newblock Learning phrase representations using {RNN} encoder-decoder for
  statistical machine translation.
\newblock {\em arXiv preprint arXiv:1406.1078}, 2014.

\bibitem{mnist}
Li~Deng.
\newblock The {MNIST} database of handwritten digit images for machine learning
  research [best of the web].
\newblock {\em IEEE Signal Processing Magazine}, 29(6):141--142, 2012.

\bibitem{mitbih}
Ary~L. Goldberger, Luis A.~N. Amaral, Leon Glass, Jeffrey~M. Hausdorff,
  Plamen~Ch. Ivanov, Roger~G. Mark, Joseph~E. Mietus, George~B. Moody,
  Chung-Kang Peng, and H.~Eugene Stanley.
\newblock Physiobank, {PhysioToolkit}, and {PhysioNet}: components of a new
  research resource for complex physiologic signals.
\newblock {\em Circulation}, 101(23):e215--e220, 2000.

\bibitem{resnet}
Kaiming He, Xiangyu Zhang, Shaoqing Ren, and Jian Sun.
\newblock Deep residual learning for image recognition.
\newblock In {\em Proceedings of the IEEE conference on computer vision and
  pattern recognition}, pages 770--778, 2016.

\bibitem{lstm}
Sepp Hochreiter and J{\"u}rgen Schmidhuber.
\newblock Long short-term memory.
\newblock {\em Neural computation}, 9(8):1735--1780, 1997.

\bibitem{ecg}
Loren Kersey, Kat Lilly, and Noel Park.
\newblock {ECG} heartbeat classification: An exploratory study.
\newblock In {\em Proceedings of the Australasian Joint Conference on
  Artificial Intelligence-Workshops}, pages 20--27, 2018.

\bibitem{phasedlstm}
Daniel Neil, Michael Pfeiffer, and Shih-Chii Liu.
\newblock Phased {LSTM}: Accelerating recurrent network training for long or
  event-based sequences.
\newblock {\em arXiv preprint arXiv:1610.09513}, 2016.

\bibitem{sander2021momentum}
Michael~E Sander, Pierre Ablin, Mathieu Blondel, and Gabriel Peyr{\'e}.
\newblock Momentum residual neural networks.
\newblock {\em arXiv preprint arXiv:2102.07870}, 2021.

\bibitem{heun}
Dennis~G Zill.
\newblock {\em A first course in differential equations with modeling
  applications}.
\newblock Cengage Learning, 2012.

\end{thebibliography}



\end{document}